\documentclass[12pt]{article}

\usepackage{amsmath,amsthm, amsfonts, amssymb, amsxtra, amsopn}
\usepackage{pgfplots}
\pgfplotsset{compat=1.13}
\usepackage{pgfplotstable}
\usepackage{graphicx,grffile}
\usepackage{multirow}
\usepackage{booktabs}
\usepackage{listings}
\usepackage{cmap}
\usepackage{colortbl}
\usepackage{adjustbox}
\usepackage{epsfig}

\usepackage[tableposition=top,font=small,skip=5pt]{caption}
\usepackage{subcaption}
\usepackage{makecell} 

\usepackage[explicit]{titlesec}

\PassOptionsToPackage{hyphens}{url}

\usepackage{hyperref}
\hypersetup{colorlinks=true,linkcolor=black,citecolor=black,urlcolor=blue,filecolor=black}
\hypersetup{pdfpagemode=UseNone,pdfstartview=}

\definecolor{darkgreen}{rgb}{0.125,0.5,0.169}

\usepackage{enumitem}
\setlist[itemize]{noitemsep, topsep=0pt}

\advance\oddsidemargin by -0.35in
\advance\textwidth by 0.7in

\advance\topmargin by -0.45in
\advance\textheight by 0.9in

\long\def\symbolfootnotetext[#1]#2{\begingroup%
\def\thefootnote{\fnsymbol{footnote}}\footnotetext[#1]{#2}\endgroup}

\newcommand\dunderline[3][-1pt]{{%
  \sbox0{#3}%
  \ooalign{\copy0\cr\rule[\dimexpr#1-#2\relax]{\wd0}{#2}}}}
\def\uuu{\kern-1pt\dunderline{0.75pt}{\phantom{M}}}

\DeclareMathOperator{\dc}{data center}
\DeclareMathOperator{\tot}{total}
\DeclareMathOperator{\gpu}{gpu}
\DeclareMathOperator{\cpu}{cpu}
\DeclareMathOperator{\ram}{ram}

\def\zz{\phantom{0}}

\title{Energy Considerations for Large Pretrained Neural Networks}

\author{Leo Mei\footnotemark[1]\ \ \ 
Mark Stamp\footnotemark[1]\,\,\footnotemark[2]}

\begin{document}

\symbolfootnotetext[1]{Department of Computer Science, San Jose State University}
\symbolfootnotetext[2]{mark.stamp$@$sjsu.edu}

\maketitle

\abstract
Increasingly complex neural network architectures 
have achieved phenomenal performance. However, these
complex models require massive computational resources that consume substantial 
amounts of electricity, which highlights 
the potential environmental impact of such models. Previous studies have 
demonstrated that substantial redundancies exist in large pre-trained models. 
However, previous work has primarily focused on compressing models
while retaining comparable model
performance, and the direct impact on electricity consumption
appears to have received relatively little attention. By quantifying 
the energy usage associated with both uncompressed and compressed models, we 
investigate compression as a means of reducing electricity consumption. 
We consider nine different pre-trained models, 
ranging in size from 8M parameters to 138M parameters. To establish a baseline, 
we first train each model without compression and record the electricity usage
and time required during training, along with other relevant statistics. 
We then apply three compression techniques:
Steganographic capacity reduction, pruning, and low-rank factorization. 
In each of the resulting cases, we again measure the electricity usage, 
training time, model accuracy, and so on. 
We find that pruning and low-rank factorization offer no
significant improvements with respect to energy usage or other 
related statistics, while steganographic capacity reduction 
provides major benefits in almost every case. 
We discuss the significance of these findings.

\section{Introduction}\label{ch:intro}

Deep Neural Networks (DNN) have become predominant and ubiquitous in recent years~\cite{Balas_2019}, 
facilitated by advancement in hardware and training technology. Such advances have also driven the 
exponential growth of DNNs in size~\cite{Xu_Ding_Hu_Niemier_Cong_Hu_Shi_2018}.  For example, 
Krizhevshy et al.~\cite{Krizhevsky_Sutskever_Hinton_2017} made a groundbreaking achievement in 
the~2012 ImageNet Challenge using a network that contains five convolutional layers and three fully 
connected layers and a total of~60 million parameters. However, by~2020
the Large Language Model (LLM) GPT-3 had more than~175 billion 
parameters~\cite{Brown_Mann_Ryder_Subbiah_Kaplan_Dhariwal_Neelakantan_Shyam_Sastry_Askell_etal._2020}.
As the demand for advanced AI models capable of handling complex tasks increases, networks are likely to 
grow even larger, with higher parameter counts and deeper topologies. Larger models require more 
computational power to train, and hence consume more energy.

Substantial energy consumption is a known problem associated with training and deploying large 
DNNs~\cite{Atallah_Banda_Banda_Roeck_2023,Hu_Chu_Pei_Liu_Bian_2021,Thompson}. 
Krizhevshy et al.~\cite{Krizhevsky_Sutskever_Hinton_2017} spent two to three days training 
the aforementioned model on the ImageNet dataset with an Nvidia K40 GPU, 
while GPT-3 required weeks or even months to train, and required hundreds of advanced
GPUs~\cite{Brown_Mann_Ryder_Subbiah_Kaplan_Dhariwal_Neelakantan_Shyam_Sastry_Askell_etal._2020}. 

Table~\ref{tab:energy_consumption_lnn} shows the energy consumption associated with the training of an LLM 
as compared to the average annual energy consumption of a US household in~2022.  During the training process, 
GPT-3 consumed approximately~1287 megawatt hours (MWh), while another large language model, 
Gopher, consumed approximately~1066 MWh. In contrast, the average annual energy consumption of 
a U.S. household in 2022 was estimated to be around~11 MWh. This comparison reveals the significant 
energy consumption of large neural network training. 

\begin{table}[!htb]
    \centering
    \def\a{\phantom{$\approx$ }}
    \def\c{\phantom{,}}
    \def\z{\phantom{\hbox{0}}}
    \caption{Energy consumption required to train large DNNs~\cite{Clark_Perrault}}
    \label{tab:energy_consumption_lnn}
    \begin{adjustbox}{scale=0.85}
    \begin{tabular}{lc}
        \toprule
        \multirow{2}{*}{\textbf{Neural network}} & \textbf{Energy consumption} \\
        		& \textbf{per model training (MWh)} \\
        \midrule
        GPT-4   & $\approx$ 3,500 \\
        GPT-3   & \a 1,287 \\
        Gopher  & \a 1,066 \\
        BLOOM   & \a \z\c433 \\
        OPT     & \a \z\c324 \\
        \midrule
        U.S. household average annual & \a \z\c\z11 \\
        \bottomrule
    \end{tabular}
    \end{adjustbox}
\end{table}

Thompson, et al.~\cite{Thompson} emphasize concerns about diminishing returns of DNNs as 
compared to energy consumption. They reviewed the decreasing gains in accuracy improvements, 
noting that the annual improvement in the top-performing ImageNet models could drop to as little as~5\%\ by~2025. 
However, the substantial computing resources and energy required to 
train such advanced systems could result in consumption equivalent to that of New York City for 
one month. To sustain improvement in model performance, researchers are developing increasingly 
larger and deeper models whose size is growing 
exponentially~\cite{Bianco_Cadene_Celona_Napoletano_2018,Xu_Ding_Hu_Niemier_Cong_Hu_Shi_2018}. 
This growth of models results in a substantial increase in the energy consumption required for training.

DNNs are also being employed in various smart applications, ranging from autonomous robots 
to personal smart wearable devices, such as smartwatches and fitness trackers. These devices 
typically require real-time computation. For example, a self-driving car must process real-time video to 
detect obstacles on the road to avoid collisions, which requires its DNN to produce results efficiently and 
effectively. However, these devices also often operate under strict constraints, including limited 
computational power and battery capacity. If the size of DNN models continues to grow, 
the time they take to process inputs will likely increase, resulting in more energy consumption 
and shorter battery life for mobile applications~\cite{Liu_Lin_Zhou_Nan_Liu_Du_2018}. 
As a result, the deployment of larg DNNs on resource-constrained devices presents a 
significant challenge, due to substantial energy consumption.

Previous studies have shown that DNNs have remarkable redundancies and that these 
redundancies can be reduced without compromising model 
performance~\cite{AJOPPSSZS_short, Bucilua_Caruana_Niculescu-Mizil_2006, Cheng_Wang_Zhou_Zhang_2018}. 
One of the most effective redundancy-reduction techniques is compression, which can reduce the size and 
computational requirements of DNNs~\cite{Bucilua_Caruana_Niculescu-Mizil_2006}. Model compression offers the 
advantage of reducing the storage requirement and memory footprint, making DNNs more efficient 
and practical to deploy on resource-constrained devices, and to transmit over the network to edge devices. 
However, its effect on training time and computational costs remains uncertain~\cite{Qin_Ren_Yu_Wang_Gao_Zheng_Feng_Fang_Wang_2018}. 
However, compression may reduce model performance, and hence there is no guarantee that a 
compressed model will achieve comparable performance as an uncompressed model
trained for the same number of epochs, and a compressed models may require more training epochs 
to achieve equivalent performance as an uncompressed model. 
Thus, a compressed model may not result in lower computational costs and energy consumption. 
In this research, we investigate whether compressed models consume less electricity and 
thereby determine the effectiveness of compression as a strategy to reduce the environmental impact
of DNNs. 

We initially train models without compression and record electricity usage, which serves to establish a 
baseline for comparison. Then, we consider three compression techniques. First, we determine the 
steganographic capacity of models, and reduce the size of each model by quantizing 
the parameters to lower precision~\cite{AJOPPSSZS_short, Yang_Shen_Xing_Tian_Li_Deng_Huang_Hua_quantitize}. 
Second, we prune models to remove unnecessary parameters~\cite{Reed_pruning}. Third, 
we employ low-rank factorization, that is, we find the minimum rank into which a weight matrix 
can be decomposed without degrading the performance, and we compress models by reducing 
the dimensionality accordingly~\cite{Sainath_Kingsbury_Sindhwani_Arisoy_Ramabhadran_lora}. 
In each case, during training, we measure and record the electricity usage of each model, comparing 
energy consumption and environmental impact against model performance, in terms of classification accuracy, 
training time, compression ratio, and number of training rounds needed to achieve a given level of performance.

The remainder of this paper is organized as follows. In Section~\ref{ch:background}, 
we discuss relevant background topics. Section~\ref{ch:expSetup} provides an overview of the dataset 
used in our experiments, the experimental setup, and the performance metrics used to evaluate our
results. Our experimental results are presented and discussed in Section~\ref{ch:expResult}. 
Finally, we summarize this research and discuss potential avenues for future work in 
Section~\ref{ch:conclusion}.

\section{Background}\label{ch:background}

In this section, we provide details on relevant background topics used in this study. 
We begin by describing the various models used, highlighting their architectures and 
training parameters. Next, we discuss various model compression techniques, and 
we outline the procedure for calculating the energy consumption of training these models. 
This section also includes details on the dataset and performance metrics
used in our experiments.

\subsection{Deep Neural Network Models}\label{sec:DNNs}

Deep Neural Networks (DNN) are neural networks with many layers, including hidden layers.
DNNs are the primary architectures used in Deep Learning (DL), as such models have the 
capability of automatically learning hierarchical representations of data~\cite{LeCun_Bengio_Hinton_2015}. 
DNNs are designed to mimic the human brain, which contains numerous interconnecting 
neurons that process, transmit, and capture important information. Because DNNs
can extract massive volumes of information from raw data without intense feature engineering, 
DNNs are capable of learning patterns and relationship from large datasets. This learning process 
during training, where internal weights and parameters are modified, with the goal of 
optimizing performance for a specific task. DNNs often attain higher performance levels, 
as compared to classic Machine Learning (ML) models.

In recent years, DNNs have achieved phenomenal success and breakthroughs in various 
fields~\cite{Gallifant_2024}. For example, in computer vision, convolutional neural networks (CNN) have 
resulted in advances in object detection, medical image diagnostics, and image classification. 
In Reinforcement Learning (RL), DNNs have been used by AI agents to learn optimal strategies through 
trial and error, resulting in autonomous vehicles and robots. Similarly, deep transformer-based models,
such as ChatGPT, have revolutionized Natural Language Processing (NLP), 
making conversational AI, text generation, summarization, and translation accessible.

DNNs can be broadly categorized into two types based on their learning paradigm, namely, 
supervised learning and unsupervised learning~\cite{LeCun_Bengio_Hinton_2015}. These learning 
approaches define how models process input data for training. Supervised learning trains a DNN on 
labeled data, where each data point in the training set must include a class label. In supervised learning,
training consists of minimizing errors between model predictions and ground truth (as specified by the labels). 
In contrast, unsupervised learning does not involve labeled data, and the goal is
to learn patterns and relationships from a dataset. Unsupervised learning is commonly used for clustering, 
dimensionality reduction, and anomaly detection. 

Pre-trained models refer to machine learning neural network models that have been previously trained on 
datasets. These models can be reused for similar tasks immediately without further full-scale training. 
For example, a model pre-trained on millions of images can be fine-tuned to classify specific types of 
object with much less data~\cite{Thompson_Schonwiesner_Bengio_Willett_2019}. This approach is 
especially useful in deep learning, where training from scratch consumes a great deal of time and 
computational resources. In this study, we consider pre-trained models and fine-tune their performance 
for our specific task while simultaneously tracking the training costs involved.
Specifically, we train models to classify images from a large benchmark dataset 
containing~1000 different classes. 

For our energy experiments, we consider the 
following classes of pre-trained models:
AlexNet (Alex Krizhevsky Network), 
ResNet (Residual Network), 
Inception, 
DenseNet (Densely Connected Convolutional Network), 
VGG (Visual Geometry Group Network),
and ConvNeXt (Convolutional Neural Network Next-Generation).
Table~\ref{tab:models_charac} lists relevant characteristics of these models.

\begin{table}[!htb]
    \centering
    \def\z{\phantom{0}}
    \caption{Models tested in this paper}\label{tab:models_charac}
    \begin{adjustbox}{scale=0.85}
    \begin{tabular}{lccc}
        \toprule
        \multirow{2}{*}{\textbf{Model}} & \multirow{2}{*}{\textbf{Type}} 
        		& \textbf{Total} & \multirow{2}{*}{\textbf{Depth}} \\
		&& \textbf{parameters} \\
        \midrule
        AlexNet & CNN & \z60.0M & \z\z8 \\
        ResNet18 & CNN & \z11.7M & \z18 \\
        ResNet34 & CNN & \z21.8M & \z34 \\
        ResNet50 & CNN & \z25.6M & \z50 \\
        ResNet101 & CNN & \z44.5M & 101 \\
        InceptionV3 & CNN & \z27.1M & \z42 \\
        DenseNet121 & CNN & \z\z8.0M & 121 \\
        VGG16 & CNN, fully connected & 138.0M & \z16 \\
        ConvNeXt-Base & CNN & \z88.0M & \z89 \\
        \bottomrule
    \end{tabular}
    \end{adjustbox}
\end{table}

\subsubsection{Overview of AlexNet}

AlexNet~\cite{Krizhevsky_Sutskever_Hinton_2017} achieved a significant breakthrough in deep learning by 
winning the~2012 ImageNet Large Scale Visual Recognition Challenge (ILSVRC). 
AlexNet is a deep convolutional 
neural network, and in the ILSVRC competition it 
significantly outperforming classic ML models. This achievement demonstrated the power 
of deep learning for the large-scale image classification task, and it triggered a shift of focus 
from feature engineering to automated DL architectures.

AlexNet has eight layers, consisting of five convolutional layers for feature extraction and 
three fully connected layers for classification. It uses ReLU (Rectified Linear Unit) activation 
instead of sigmoid or tanh functions, which helps to mitigate the vanishing gradient problem. 
AlexNet also introduced dropout regularization, which is a powerful technique to reduce overfitting. 
By incorporating GPU acceleration for faster training, it also made efficient deep learning on 
a large dataset feasible.

\subsubsection{Overview of ResNet}

ResNet~\cite{He_Zhang_Ren_Sun_2016} revolutionized deep learning by mitigating the 
vanishing gradient problem. The mechanism ResNet gets its name from its use of residual 
connections, where the model can skip layers during training, which serves to improve gradient flow. 
This innovation makes it feasible to train very deep neural networks.

The ResNet family includes five variants: ResNet18, ResNet34, ResNet50, ResNet101, and ResNet152.
All of these versions include initial~$7\times 7$ convolutional layers with pooling to expand the receptive 
field and residual blocks with skip connections to learn features, while improving gradient flow. 
In this study, we consider ResNet18, ResNet34, ResNet50, and ResNet101.

\subsubsection{Overview of Inception}

Inception~\cite{Szegedy_Wei_Liu_Yangqing_Jia_Sermanet_Reed_Anguelov_Erhan_Vanhoucke_Rabinovich_2015} 
introduced the so-called Inception module, which consists of~$1\times 1$, $3\times 3$, and~$5\times 5$ convolutions 
in parallel. This enables multiscale feature extraction, thereby capturing fine and coarse details. 
Compared to other advanced CNN arachitectures, Inception has a relatively modest number of parameters, 
but in some cases, it achieves higher accuracy than more complex models. As a result, Inception can reduce 
computational costs.

The Inception family includes four variants: InceptionV1 (formerly known as GoogLeNet), 
InceptionV2, InceptionV3, and InceptionV4. All of these versions include initial convolution 
and pooling layers to create diverse feature representations, inception modules to capture fine 
and coarse feature details, and auxiliary classifiers to improve gradient flow. 
In this study, we consider InceptionV3.

\subsubsection{Overview of DenseNet}

DenseNet~\cite{Huang_Liu_Van_Der_Maaten_Weinberger_2017} improves feature reuse and gradient 
flow by connecting each layer to every other layer within a block. While traditional CNN layers only receive 
input from the previous layer, DenseNet allows reuse of features throughout the network. 
Reuse of features provides for more efficient use of parameters.

The DenseNet family includes four variants: 
DenseNet121, DenseNet169, DenseNet201, and DenseNet264. 
All of these variants consist of initial~$7\times 7$ convolutional layers with pooling to expand the receptive field;
dense blocks to enhance feature extraction and transition (bottleneck) layers with pointwise~$1\times 1$ 
convolutions and~$2\times 2$ pooling, to compress and downscale feature maps and thereby reduce model 
complexity and spatial dimensions; and direct connections between layers, which help preserve gradient 
flow and alleviate the vanish gradient. In this study, we consider DenseNet121.

\subsubsection{Overview of VGG}

VGG~\cite{Simonyan_Zisserman_2015} models employ~$3\times 3$ convolutional filters uniformly throughout 
the network, with stride one and padding one to capture spatial information. In addition, 
max-pooling layers with a~$2\times 2$ kernel are used to reduce dimensionality. VGG models are
known for their simplicity and effectiveness. By stacking multiple small filters together, 
VGG increases the receptive field while maintaining computational efficiency. 
In addition, the small filter size allows the model to capture fine-grained spatial features.

The VGG family includes four variants: VGG11, VGG13, VGG16, and VGG19. 
In this study, we consider VGG16.

\subsubsection{Overview of ConvNeXt}

ConvNeXt~\cite{Liu_Mao_Wu_Feichtenhofer_Darrell_Xie_2022} is a modern CNN architecture that 
incorporates design elements from transformer architectures. Its performance demonstrates that 
CNNs can be competitive with advanced transformer models in image recognition tasks, while preserving 
computational efficiency. This has significant implications for real-world applications such as 
autonomous vehicles, medical image diagnostics, and mobile AI.

The ConvNeXt family includes four variants: ConvNeXt-Tiny, ConvNeXt-Small, ConvNeXt-Base, and ConvNeXt-Large. 
All of these variants include depthwise~$7\times 7$ convolutions to expand the receptive field, 
layer normalization for stable training, an inverted bottleneck with pointwise~$1\times 1$ convolutions, 
GELU activation for improved non-linearity, and a residual connection to enhance gradient flow. 
In this study, we consider ConvNeXt-Base, which we subsequently refer to as simply ConvNeXt. 

\subsection{Compression Techniques}\label{sec:CT}

Training neural network models often demands substantial computational power, 
which results in high energy consumption. Recently, concerns about the environmental 
impact of training deep neural networks have grown as large language models have 
gained popularity~\cite{Clark_Perrault, Thompson}. Previous work has shown that 
neural networks can be compressed without significantly compromising their 
performance~\cite{Long_Ben_Liu_2019,MPMFrasca_2023}. In this context, comparing the energy 
consumption of training full, uncompressed models to that of compressed models is relevant. 
Compressed models typically have few parameters or use lower precision weights.  As a result, 
they may have reduced computational demands and shortened training time, 
resulting in lower energy usage and a smaller carbon footprint. Next, we discuss introduce
the compression techniques considered in this paper, namely, steganographic capacity, 
pruning, and low-rank factorization.

\subsubsection{Steganographic Capacity}

The word ``steganography'' originates from the Greek words ``steganos'' (covered) and ``graphein'' (writing), 
which together imply hiding information in writing~\cite{Johnson_Duric_Jajodia_Memon_2001}. 
In modern practice, steganography refers to hiding information in another medium, 
such as text, images, video, or audio~\cite{Bender_Gruhl_Morimoto_Lu_1996}. While \hbox{cryptography} 
focuses on encrypting information by converting it to a form that is unreadable without a key,
steganography aims to conceal the existence of the information itself in another medium,
without changing the properties of the carrier medium~\cite{Katzenbeisser_Petitcolas_2000}. 
For example, 
in an uncompressed RGB format, a pixel uses one byte for each of its R (red), G (green), and B (blue)
components. As a result, there are~$2^{24}$ color combinations per pixel, and many of these combinations are 
not distinguished by the human eye, resulting in considerable redundancy. Thus, the low-order bits 
of each RGB byte can be used to embed information, without affecting the perceived image.

Pretrained DNNs have been shown to exhibit a high steganographic capacity~\cite{AJOPPSSZS_short}. 
Specifically, a substantial number of the low-order bits of the trained parameters can be overwritten 
without degrading the performance of such models. The authors of~\cite{AJOPPSSZS_short}
experimented with various models, including Linear Regression (LR), Support Vector Machine (SVM), 
Multilayer Perceptron (MLP), Convolutional Neural Network (CNN), Long Short-Term Memory (LSTM), 
VGG16, DenseNet121, InceptionV3, Xception, and ACGAN. They found that the per-weight steganographic
capacity of these models ranged from~19 to~27 bits per 32-bit weight, implying a minimum compression
ratio of more than~0.59\ to a maximum of more than~0.84.

In this paper, we determine the steganographic capacity of each model under consideration,
and we reduce model size by then quantizing model 
weights~\cite{Yang_Shen_Xing_Tian_Li_Deng_Huang_Hua_quantitize}. 
We use a threshold of a decrease in accuracy of no more than~1\%.
In this way, we are able to compress a model without significantly degrading the performance.  

\subsubsection{Pruning}

Model pruning is a compression technique that is used to reduce the size of a neural network 
by removing redundant weights or neurons. This concept was introduced 
in~\cite{han2015deep_compression} and is based on the insight that most weights in 
DNNs are redundant and can therefore be eliminated with little loss of performance. 
The pruning process begins by training a model normally to achieve high performance. 
Then, the contribution of each weight to the output is determined. By removing weights or 
neurons that contribute less than a specified threshold, the impact on model performance is
minimal~\cite{Reed_pruning}. After pruning, remaining weights are fine-tuned to compensate 
for performance loss caused by the pruning. 

Pruning reduces the number of weights, which thereby reduces the number of multiplication 
and addition operations. 
This results in fewer computations during training, which reduces overall training time.
In addition, pruning is particularly useful for deploying models on devices with limited computational 
resources, as it also reduces memory access by minimizing data read/write operations.
In this study, we gradually increase the pruning ratio for each model to determine the optimal 
pruning percentage, which we define as causing no more than a~1\%\ drop in accuracy. 

\subsubsection{Low-Rank Factorization}\label{sub:lora}

Low-rank factorization is based on a mathematical approach that approximates a 
matrix~$A$ by the product of smaller matrices with lower rank~\cite{Eckart_Young_1936}. 
The primary goal is to represent the original matrix~$A$ with fewer parameters, 
while preserving important features of the original matrix. 

The relevant factorization of~$A$ is given by
$$
A \approx U \cdot V 
$$
where~$U \in \mathbb{R}^{m \times r}$, $V \in \mathbb{R}^{r \times n}$,
and the rank satisfies~$r \ll \min(m, n)$.

A trained model can be compressed by applying low-rank factorization to its weight 
matrices~\cite{Sainath_Kingsbury_Sindhwani_Arisoy_Ramabhadran_lora}. The rank of the approximation 
used in low-rank factorization is a critical factor, as the rank determines the compression rate of the 
neural network. One common technique to minimize the rank of a matrix is 
Singular Vector Decomposition (SVD)~\cite{denton_svd}, with~$A$ factorized as
$$
    A \approx U_r \Sigma_r V_r^T
$$
where~$U_r \in \mathbb{R}^{m \times r}$, 
$\Sigma_r \in \mathbb{R}^{r \times r}$, 
$V_r \in \mathbb{R}^{n \times r}$, 
and~$r$ is the rank of the approximation.

In our experiments, $A$ is the weight matrix of a trained model, and
we dynamically adjust~$r$ to find an approximation of the weight matrices 
that results in minimal loss of accuracy for the models. Specifically, for each layer, 
we calculate the minimum rank that causes a decrease in accuracy less than~1\%. 

\subsection{Calculating Energy Consumption}

The power consumed during training includes the total consumption of the 
CPU, GPU, and RAM. Energy consumption is calculated in kilowatt-hours (kWh) as
$$
    P_{\tot}=P_{\gpu}+P_{\cpu}+P_{\ram}
$$
This measurement ensures that all components that contribute to energy use 
are taken into account.

Our models are trained on Kaggle, which is hosted in a data center, and hence we cannot
directly measure energy consumption.
Therefore, we consider the Power Usage Effectiveness (PUE)~\cite{Avelar_Azevedo_French_2014},
which quantifies the efficiency of a computer data center facility by comparing the total energy used by 
the facility with the energy delivered to the computing equipment. The PUE is calculated as
$$
    \mbox{PUE} = \frac{\mbox{Total Facility Energy}}{\mbox{IT Equipment Energy}} = 1 
	+ \frac{\mbox{Non IT Facility Energy}}{\mbox{IT Equipment Energy}}
$$
PUE provides insight into how efficiently a data center utilizes energy, specifically highlighting 
the balance between the energy used by the computing equipment and the energy expended 
on cooling and other overhead tasks. According to Uptime Institute~\cite{dbizo@uptimeinstitute.com_2023}, 
the average PUE for a data center in 2023 is~1.58. Therefore, the formula to calculate the total power 
consumed by the data center is
$$
    P_{\dc}=1.58 \cdot P_{\tot}
$$

\subsection{Dataset and Performance Metrics}\label{ch:expSetup}

We use ImageNet1K~\cite{Deng_Dong_Socher_Li_Kai_Li_Li_Fei-Fei_2009} for all of our experiments.
It is a widely used benchmark dataset for image-related machine learning tasks,
consisting of~1,000 classes with~1,281,167 training images, 50,000 validation images, and~100,000 test images. 
Each class contains annotated images of real-life objects that correspond to an
object description, making it an ideal dataset for evaluating DNNs. Table~\ref{tab:imagenet_classes} 
lists the first~11 classes of ImageNet1K.

\begin{table}[!htb]
    \centering
    \def\z{\phantom{0}}
    \caption{First 11 classes of ImageNet1K}\label{tab:imagenet_classes}
    \begin{adjustbox}{scale=0.85}
        \begin{tabular}{cl}
            \toprule
            \textbf{Class} & \textbf{Images} \\
            \midrule
            \z0 & tench \\
            \z1 & goldfish \\
            \z2 & great white shark, white shark, man-eating shark \\
            \z3 & tiger shark \\
            \z4 & hammerhead, hammerhead shark \\
            \z5 & electric ray, crampfish, numbfish, torpedo \\
            \z6 & stingray \\
            \z7 & cock \\
            \z8 & hen \\
            \z9 & ostrich \\
            10 & brambling \\
            \bottomrule
        \end{tabular}
    \end{adjustbox}
\end{table}


Evaluation metrics quantify the performance of a model on a specific task,
providing guidance on how to align outcomes with real-world 
needs~\cite{Rainio_Teuho_Klen_2024}. 
In this study, we consider the following metrics.
\begin{description}
\item[Accuracy] is the ratio of correctly predicted samples to the total number of samples classified.
\item[Energy consumption] is the total energy used for model training. This is the most essential 
and intuitive metric to evaluate the efficiency of the compression techniques considered in this paper.
\item[Training duration] is the execution time required to train a model, 
excluding model initialization and loading time. 
It is highly correlated to energy consumption.
\item[Number of epochs] is the number of iterations through the training data required
to reach a particular level of performance. This measure enables us to compare the efficiency 
of training a compressed model, as compared to the original model.
\end{description}
Obviously, the higher the accuracy, the better. For each of the other three metrics listed above,
lower is better, with energy consumption being the focus of this research.


As mentioned above, our experiment platform is Kaggle. 
Kaggle's cloud-based hardware system consists of an Intel Xeon 2.00 GHz CPU, 32GB RAM, 
and two NVIDIA~T4 GPUs, each with~16GB of memory.
We train our models using Jupyter Notebook on Kaggle, leveraging PyTorch 
for building and training the models, while CuDNN and CUDA are used to 
accelerate GPU computations.

\section{Experiments and Results}\label{ch:expResult}

In this section, we first discuss the hyperparameters used when training the models under
consideration. Then we give our experimental results
for each of the three compression techniques 
discussed in Section~\ref{sec:CT}, namely, steganographic capacity reduction, pruning,
and low-rank factorization.
In each case, we test all of the pretrained models introduced in Section~\ref{sec:DNNs},
namely,
AlexNet,
ResNet18,
ResNet34,
ResNet50,
ResNet101,
InceptionV3,
DenseNet121,
VGG16, and
\hbox{ConvNeXt}.
We conclude this section with a discussion of the results or our experiments.

\subsection{Hyperparameters}

As discussed above, our experiment aims to determine the minimum energy consumption 
required for a compressed model to achieve reasonable performance, as compared to the original model. 
Unlike the typical goal of pursuing the best accuracy, our goal is to investigate whether compressed 
models---that achieve comparable accuracy to the original model---consume less electricity and 
thereby determine whether compression is an effective strategy to reduce the environmental impact. 
Therefore, we need to ensure uniform training parameters across all models throughout the training process. 
This consistency eliminates potential confounding variables that would be introduced by varying 
hyperparameters, enabling us to focus on the impact of compression on 
energy consumption. 

SGD with momentum is commonly used for image classification tasks and has been proven to be effective and 
to generalize well when training CNNs~\cite{wilson_2017}. We choose a common value of~0.9 for momentum, 
as this value is used in several standard models, including AlexNet, ResNet, and VGG. A momentum value 
of~0.9 can generally speed up convergence, while typically avoiding oscillations. Because ImageNet1K is a 
relatively large dataset, we select a learning rate of~0.01 to further accelerate training. This learning rate 
was used in the original VGG paper~\cite{Simonyan_Zisserman_2015}. To avoid overfitting, we pick a 
weight decay coefficient of~0.0001, which is generally considered to be 
a reasonable value for image classification using CNNs with SGD and momentum~\cite{decay}. 
These training hyperparameters discussed here 
are summarized in Table~\ref{tab:training_params}. 

\begin{table}[!htb]
    \centering
    \caption{Selected hyperparameters}\label{tab:training_params}
    \begin{adjustbox}{scale=0.85}
    \begin{tabular}{ll}
        \toprule
        \textbf{Parameter} & \textbf{Value} \\
        \midrule
        Optimizer & SGD \\
        Learning Rate & 0.01\zz\zz \\
        Weight Decay & 0.0001 \\
        Momentum & 0.9\zz\zz\zz \\
        \bottomrule
    \end{tabular}
    \end{adjustbox}
\end{table}

We also employ early stopping to avoid overfitting. 
Specifically, we stop the training process if there is no improvement of~5\%\ or more in validation 
loss for three consecutive epochs. In ImageNet1K training, we find that validation loss 
typically exhibits plateau patterns, at which point further training will result in diminished 
results. These parameters related to early stopping 
are summarized in Table~\ref{tab:training_early}.

\begin{table}[!htb]
    \centering
    \caption{Early stopping}\label{tab:training_early}
    \begin{adjustbox}{scale=0.85}
    \begin{tabular}{lll}
        \toprule
        \textbf{Technique} & \textbf{Value} & \textbf{Description} \\
        \midrule
        Patience & 3 & Epochs to wait for loss improvement before stopping \\
        Min delta & 0.05 & Minimum improvement in validation loss \\
        \bottomrule
    \end{tabular}
    \end{adjustbox}
\end{table}

\subsection{Steganographic Capacity Experiments}\label{sub:stego}

In this subsection, we determine the steganographic capacity of each of the nine models 
discussed in Section~\ref{sec:DNNs}. Specifically, each trained model weight is represented in 
32-bit floating-point format, and we incrementally overwrite the low-order bits with~0 
until the model accuracy drops by more than~1\%, as compared to the baseline accuracy. 
That is, if overwriting the lowest~$n$ bits of all model weights causes a drop in model accuracy
of less than~1\%, but overwriting~$n+1$ bits causes the accuracy to drop by more than~1\%,
then the model has a steganographic capacity of~$n$ bits per weight. The steganographic capacity 
determines the quantization level, which in turn defines a compression ratio of a model. 
For each model, we graph the model accuracies as a function of the number of bits overwritten.


\begin{figure}[!htb]
    \centering\advance\tabcolsep by -5pt
    \begin{tabular}{ccc}
    \includegraphics[width=0.285\textwidth]{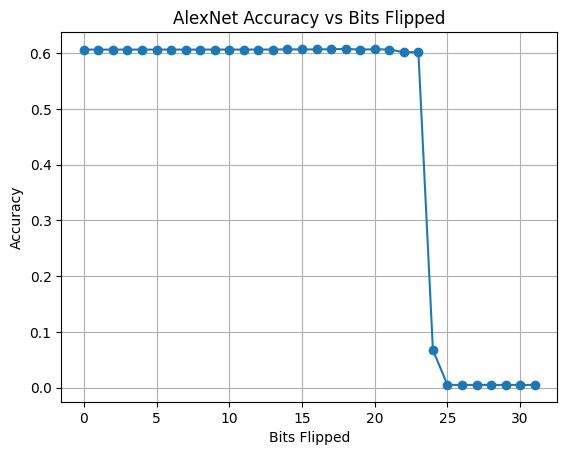}
    &
    \includegraphics[width=0.285\textwidth]{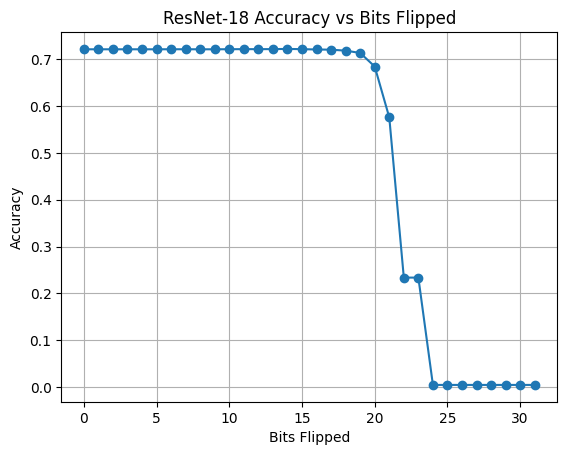}
    &
    \includegraphics[width=0.285\textwidth]{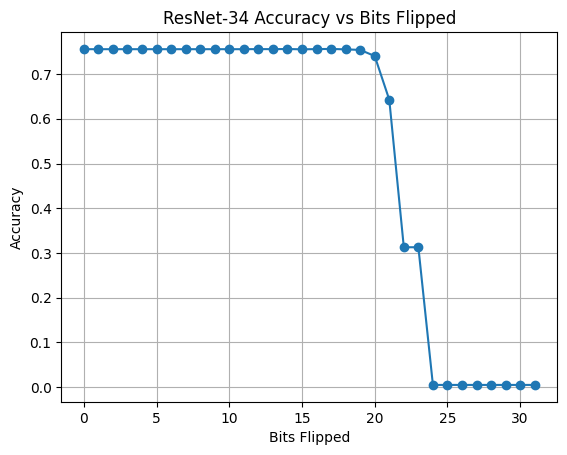}
    \\ \\[-3.5ex]
    \adjustbox{scale=0.85}{(a) AlexNet}
    &
    \adjustbox{scale=0.85}{(b) ResNet18}
    &
    \adjustbox{scale=0.85}{(c) ResNet34}
    \\ \\[-1.65ex]
    \includegraphics[width=0.285\textwidth]{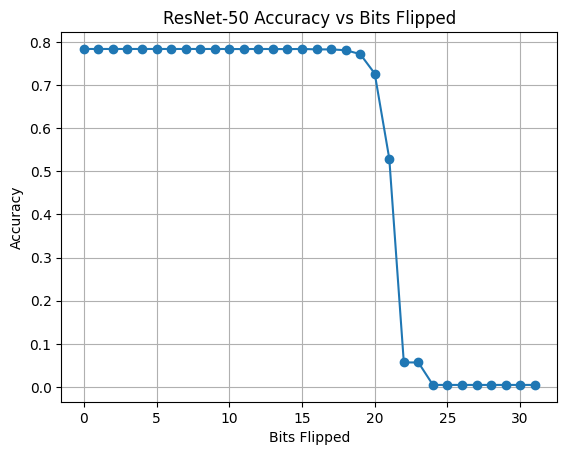}
    &
    \includegraphics[width=0.285\textwidth]{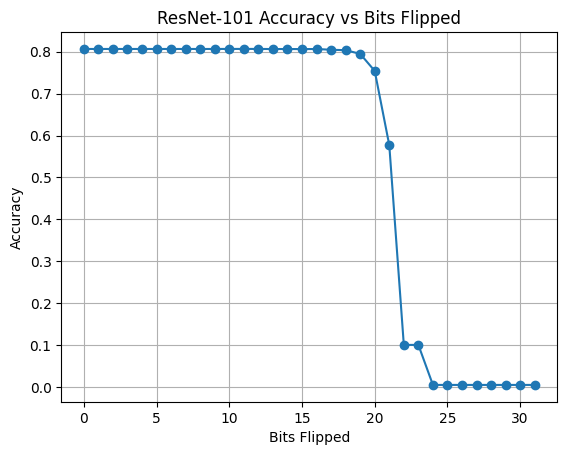}
    &
    \includegraphics[width=0.285\textwidth]{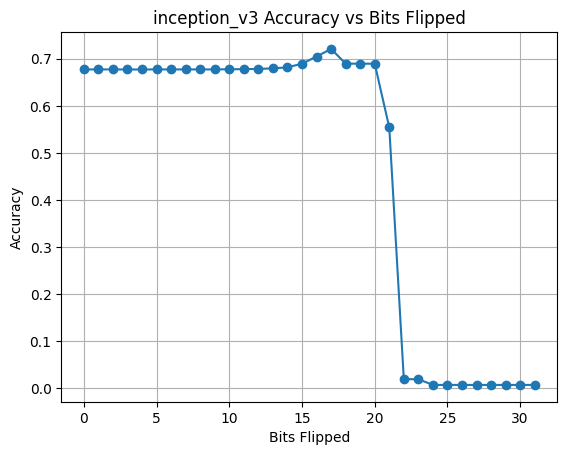}
    \\ \\[-3.5ex]
    \adjustbox{scale=0.85}{(d) ResNet50}
    &
    \adjustbox{scale=0.85}{(e) ResNet101}
    &
    \adjustbox{scale=0.85}{(f) InceptionV3}
    \\ \\[-1.65ex]
    \includegraphics[width=0.285\textwidth]{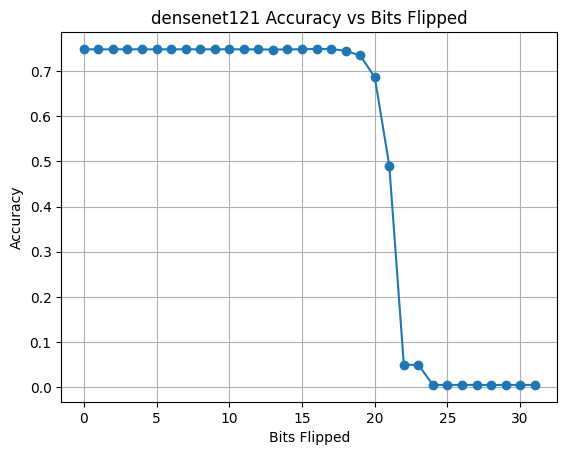}
    &
    \includegraphics[width=0.285\textwidth]{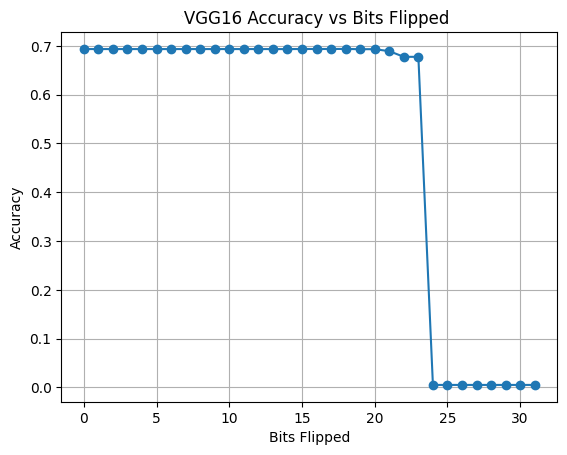}
    &
    \includegraphics[width=0.285\textwidth]{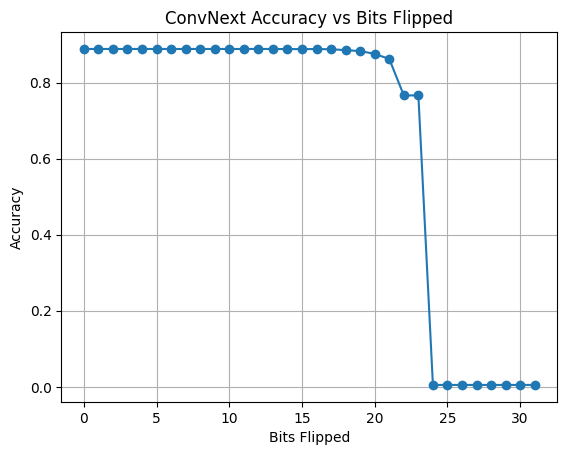}
    \\ \\[-3.5ex]
    \adjustbox{scale=0.85}{(g) DensNet121}
    &
    \adjustbox{scale=0.85}{(h) VGG16}
    &
    \adjustbox{scale=0.85}{(i) ConvNeXt}
    \end{tabular}    
    \caption{Steganographic capacity results}\label{fig:ss}
\end{figure}

\begin{description}

\item[AlexNet]---
The baseline accuracy of AlexNet on the ImageNet1K subset is 0.6062. 
There is no drop in model accuracy when we overwrite the lower~24 bits, 
but overwriting the lower~25 bits causes a large drop, as can be seen in 
Figure~\ref{fig:ss}(a). Therefore, we determine that AlexNet has~24 bits of 
steganographic capacity, and hence~8 bits is sufficient to store
each weights, yielding a compression ratio of~0.75.

\item[ResNet18]---
For ResNet18, we observe from Figure~\ref{fig:ss}(b)
that the model exhibits~19 bits of steganographic capacity. 
Consequently, the compression ratio of this model is~0.5938.

\item[ResNet34]---
Figure~\ref{fig:ss}(c) shows how the ResNet34 model accuracy changes in response to 
the overwriting of low-order bits. At~0.7586, ResNet34 has a slightly higher baseline 
accuracy than ResNet18, but the compression ratio is identical for both at~0.5938.

\item[ResNet50]---
The results of overwriting the low-order bits of ResNet50 are shown in Figure~\ref{fig:ss}(d). 
ResNet50 has a baseline accuracy of~0.7840, which is slightly higher than ResNet34.
In addition, ResNet50 the same compression ratio of~0.5938 as both ResNet18 and ResNet34.

\item[ResNet101]---
ResNet101 has the highest accuracy of the ResNet models considered, at~0.8063. 
This is not surprising, as it is the deepest architecture of our four ResNet models.
However, as shown in Figure~\ref{fig:ss}(e), the steganographic capacity per weight 
is slightly lower, since overwriting the lower~19 bits results in a drop in accuracy that
exceeds our~1\%\ threshold. Therefore, this model has~18 bits of steganographic capacity, 
which implies a compression ratio of~0.5625.

\item[InceptionV3]---
Figure~\ref{fig:ss}(f) shows that InceptionV3 begins to lose accuracy---from the baseline accuracy 
of~0.7952---when the lower~21 bits are overwritten. Therefore. the model has~20 bits of 
steganographic capacity, which translates into a compression ratio of~0.6250.

\item[DenseNet121]---
The baseline accuracy of DenseNet121 is~0.6990. As we incrementally overwrite the low-order bits, 
we find that a significant drop in accuracy occurs at bit~19; see Figure~\ref{fig:ss}(g). 
Therefore, 18 bits is the steganographic capacity, resulting in a compression ratio of~0.5625.

\item[VGG16]---
From Figure~\ref{fig:ss}(h), we observe that our VGG16 model has with a relatively low baseline 
accuracy of~0.6930, and the accuracy drop first exceeds a~1\%\ threshold when~21 low-order bits are 
overwritten. Hence, the steganographic capacity of this model is~20 bits,
resulting in a compression ratio of~0.6250.

\item[ConvNeXt]---
ConvNeXt achieves an accuracy of~0.8886 when tested on the ImageNet1K
dataset, which is impressive, given that there are~1000 classes in this dataset. 
After overwriting, the accuracy drops substantially when the low-order~20 bits are
overwritten, as shown in Figure~\ref{fig:ss}(i). As a result, we can overwrite the~19 
low-order bits of each weight, which is the steganographic capacity,
yielding a compression ratio of~0.5938 for this model.

\end{description}

The results of these steganographic capacity experiments---along with baseline accuracies---are summarized
in Table~\ref{tab:stego_rates}. We note that the accuracies vary dramatically, ranging from a high
of over~0.80 for ConvNeXt, to a low of just over~0.51 for AlexNet.

\begin{table}[!htb]
    \centering
    \caption{Summary of steganographic capacity results}\label{tab:stego_rates}
    \begin{adjustbox}{scale=0.85}
    \begin{tabular}{cccc}
        \toprule
        \multirow{2}{*}{\textbf{Model}} & \textbf{Compression} & \multicolumn{2}{c}{\textbf{Accuracy}} \\ \cline{3-4}
        		& \textbf{rate} & \textbf{Baseline} & \textbf{Compressed} \\
        \midrule
         AlexNet & 0.7500 & 0.5108 & 0.5112 \\
         ResNet18 & 0.5938 & 0.6460 & 0.6330 \\
         ResNet 34 & 0.5938 & 0.6810& 0.6760 \\
         ResNet50 & 0.5938 & 0.7260 & 0.7180 \\
         ResNet101 & 0.5625 & 0.7410 & 0.7400 \\
         InceptionV3 & 0.6250 & 0.7460 & 0.7450  \\
         DenseNet121 & 0.5625 & 0.6990 & 0.6880 \\
         VGG16 & 0.6250 & 0.6550 & 0.6660 \\ 
         ConvNeXt & 0.5938 & 0.8020 & 0.8080  \\
        \bottomrule
    \end{tabular}
    \end{adjustbox}
\end{table}

\subsection{Pruning Experiments}\label{sub:prune}

We implement a post-training pruning process with the 
help of the global unstructured pruning function in PyTorch~\cite{Pruning_Tutorial_PyTorch}. 
The function prunes the model globally by removing a specified percentage of the
least significant weights over the entire models. 
This is accomplished using global unstructured pruning, where all weights from the selected layers are 
treated as a single pool. These weights are ranked by magnitude and the smallest are pruned, regardless 
of which layer they belong to. This approach results in varying pruning percentages across different layers, 
depending on the distribution of small-magnitude weights---layers with more low-magnitude weights 
are pruned more heavily, while layers with more high-magnitude weights are pruned less. 
Pruning by magnitude is widely used, as smaller weights contribute less to a
model's output. 
 
One of the important input parameters of the pruning function is the pruning method. 
In our study, we employ ``L1 unstructured pruning''~\cite{Kumar_Shaikh_Li_Bilal_Yin_2021}. 
Here, ``unstructured'' means that the pruning occurs at the level of single weights, regardless 
of their position in the model, while~L1 refers to the~L1 norm, that is, we
pruned the weight with the smallest absolute value. 
Another important parameter is the pruning rate. 
Analogous to the steganographic capacity 
experiments discussed above, we increase the pruning rate by~1\%\ until the 
model accuracy drops by more than~1\%.
Since pruning is performed on the test set, in some cases, a slightly higher or lower
accuracy may be achieved on the validation set.
Note that the compression rate is the ratio of removed 
weights to the total number of weights in the model. 

\begin{description}

\item[AlexNet]---
With pruning, we are able to compress AlexNet by 0.46 while 
attaining an accuracy of 0.5250. This accuracy is slightly higher than the baseline accuracy.

\item[ResNet18]---
ResNet18 consists of only 18 layers; hence, there are limited 
redundancies in the model. 
With pruning, we achieve a model compression ratio of~0.28. 
The accuracy of pruned ResNet18 is~0.6334.

\item[ResNet34]---
The depth of ResNet34 is almost double of ResNet18, which implies that it should 
contain more redundant weights. With a model compression ratio ratio of~0.35,
we see that this is indeed the case.
The pruned model produces correct predictions in~0.6744 of cases.

\item[ResNet50]---
After compressing by~0.33 via pruning, ResNet50 achieves a strong accuracy of~0.7298. 
Despite the substantial reduction in parameters, this accuracy represents a slight 
improvement, as compared to the baseline model. 

\item[ResNet101]---
ResNet101 is the deepest model that we consider from the ResNet family. 
Somewhat surprisingly, this model does not exhibit a higher pruning ratio than
the other ResNet models, 
as pruning only reduces the model size by~0.27.
The pruned model attains an accuracy of~0.7304.

\item[InceptionV3]---
Pruned InceptionV3 yields an accuracy of~0.7314 with a compression rate of~0.35. 
In this case, InceptionV3 performance remains on par with the 
baseline accuracy, with only a~1.6\%\ drop in accuracy.

\item[DenseNet121]---
Pruning DenseNet121 
achieves an accuracy of~0.6784,
which is approximately~2\%\ lower than the baseline accuracy. 
The model is compressed by~0.34. 

\item[VGG16]---
VGG 16 is the second smallest model considere, with  AlexNet being the smallest.
We are able to prune~0.61 of VGG16 weights. However, this pruning level results in
an accuracy of~0.6160, which is 
a substantial drop in model accuracy, as compared to the baseline model.

\item[ConvNeXt]---
For the ConvNeXt model, we achieve a model compression rate of~0.27. 
ConvNeXt performs well after pruning, with
an high accuracy of~0.8043 for the~1000 class 
classification experiment under consideration. 

\end{description}

Table~\ref{tab:prune_rates} summarizes the pruning rates for each of the nine models.
The range in the pruning rate per layer is large in all cases, with ResNet18 having the 
smallest range, while VGG16 has the largest range. Interestingly, ResNet18 also has
lowest overall compression rates, while VGG16 achieves the highest compression 
rate, based on pruning.

\begin{table}[!htb]
    \centering
    \caption{Summary of pruning rates}\label{tab:prune_rates}
    \begin{adjustbox}{scale=0.85}
    \begin{tabular}{ccccc}
        \toprule
        \multirow{2}{*}{\textbf{Model}} & \multicolumn{2}{c}{\textbf{Per layer rate}} & \textbf{Overall}
        	& \multirow{2}{*}{\textbf{Accuracy}} \\ \cline{2-3}
        		& \textbf{Minimum} & \textbf{Maximum} & \textbf{compression} \\
        \midrule
         AlexNet & 0.0810 & 0.4930 & 0.46 & 0.5250 \\
         ResNet18 & 0.1121 & 0.3769 & 0.28 & 0.6334 \\
         ResNet 34 & 0.1181 & 0.4445 & 0.35 & 0.6744 \\
         ResNet50 & 0.1240 & 0.4721 & 0.33 & 0.7298 \\
         ResNet101 & 0.1124 & 0.7225 & 0.27 & 0.7304 \\
         InceptionV3 & 0.0775 & 0.6086 & 0.35 & 0.7314 \\
         DenseNet121 & 0.1382 & 0.8202 & 0.34 & 0.6784 \\
         VGG16 & 0.0174 & 0.7012 & 0.61 & 0.6160 \\
         ConvNeXt & 0.1851 & 0.7811 & 0.27 & 0.8043 \\
        \bottomrule
    \end{tabular}
    \end{adjustbox}
\end{table}

\subsection{Low-Rank Factorization Experiments}\label{sub:lr}

We perform low-rank factorization on the weight matrices of each layer individually to reduce the 
computational cost while maintaining accuracy. This involves decomposing each weight matrix into 
a product of smaller matrices using Singular Value Decomposition (SVD), and preserving only the most 
significant components up to a specified rank. By lowering the effective rank, we significantly reduce the 
number of matrix operations required during training. The original weight matrix is then approximated by 
these smaller matrices with the most significant ranks. We select the rank for each layer to ensure that 
the drop in validation accuracy does not exceed~1\%. That is, for a given weight matrix, if rank~$n+1$ 
causes a loss of more than~1\%\ in the model accuracy, we then consider~$n$ to be the best rank for 
that particular layer. 

After estimating the optimal rank for each individual layer, we reassemble the model and perform training. 
The result is then used to dynamically adjust the ranks, which modifies the ranks of individual layers after 
evaluating the combined accuracy loss resulting from low-rank factorization. If there is a significant drop in 
accuracy in the compressed model, we identify layers with relatively low ranks and increase their ranks 
to preserve more information. This helps recover important details that may have been lost during compression 
and improves overall model performance. We iterate this process to maximize overall performance. 
This approach results in minimal accuracy loss across all models except ResNet18, 
which experiences a~4.5\%\ drop in accuracy, as compared to the baseline model.

\begin{description}

\item[AlexNet]---
Under our low-rank factorization approach, 
the weight matrices for the layers of AlexNet vary from a minimum rank of~56
to a maximum of~248, with an average of~153.6.  
Overall, these ranks compress the model by~0.2875, and this 
compressed version of AlexNet achieves an accuracy of~0.5240, 
which is~1.39\%\ higher than the baseline accuracy.

\item[ResNet18]---
Using our low-rank factorization technique, we can compress
ResNet18 by~0.2242. However, this low-rank factorization results in 
a substantial drop in accuracy of~4.58\%\ to~0.6002.
The ranks range from a low of~40 to a high of~344, with a
per-layer average of~141.2.

\item[ResNet34]---
ResNet34 performs better than ResNet18 after low-rank factorization, 
with an accuracy 
of~0.6890, which is marginally
better than that of the baseline case. However, the compression rate
in this case is only~0.2040, and the layer ranks range from~64
to~312, with an average of~151.78.

\item[ResNet50]---
ResNet50 correctly predicts~0.7018 of the test samples in the ImageNet1k dataset,
as compared to~0.7040 for the baseline model. 
The compression rate is~0.2856. The range in the ranks at
each layer is the widest of any of the models considered. 

\item[ResNet101]---
For the ResNet101 model, we are able achieve a compression ratio 
of~0.4346 using low-rank factorization, 
which is the highest among the ResNet models considered. 
This compressed model achieves an accuracy of~0.7256, 
which is the best among the low-rank compressed ResNet models, but it is a large drop
from the~0.8063 accuracy of the baseline model. 
Similar to ResNet50, the range in the ranks over the layers is
extremely large

\item[InceptionV3]---
The low-ranks for InceptionV3  range from~37 to~392, with an average of~104.61.
This model achieves a~0.3411 compression rate, and
an accuracy of~0.7312. The accuracy is a decrease of~1.52\%\ from 
the baseline model. 

\item[DenseNet121]---
DenseNet121 has the lowest redundancy with respect to low-rank factorization 
among all models considered, as the compression rate is only~0.1760. 
The per-layer ranks range from~32 to~356, with an average of~60.8.
The accuracy of this compressed model is~0.6896, which
is less than a~1\%\ drop, as compared to the baseline model. 

\item[VGG16]---
VGG16 achieves an accuracy of~0.6390 after we perform low-rank factorization, 
which is an improvement over the~0.6250 achieved by the uncompressed baseline model.
Low-rank factorization reduces the size of this model by~0.3449.
The minimum rank for any layer is~27, while the maximum is~356,
with an average of~191.62.

\item[ConvNeXt]---
For ConvNeXt, we have a minimum rank of~8 and a maximum of~224,
with an average rank per layer of~21.4.
This yields an overall compression rate of~0.4011, and this model
maintains an accuracy of~0.8017, which is only~0.3\%\ drop 
from the baseline.

\end{description}

Table~\ref{tab:LR_rates} summarizes our low-rank factorization results for each of the nine models.
The range in the rank per layer is large in all cases, with AlexNet having the 
smallest range, while ResNet50 has the largest range, with ResNet101 having a similarly
large range. ResNet101 also attains
the highest overall compression rate, while DenseNet121 has the lowest compression 
rate, based on low-rank factorization.

\begin{table}[!htb]
    \centering
    \caption{Summary of low-rank factorization results}\label{tab:LR_rates}
    \begin{adjustbox}{scale=0.85}
    \begin{tabular}{cccccc}
        \toprule
        \multirow{2}{*}{\textbf{Model}} & \multicolumn{3}{c}{\textbf{Rank per layer}} & \textbf{Overall}
        		& \multirow{2}{*}{\textbf{Accuracy}} \\ \cline{2-4}
        		& \textbf{Minimum} & \textbf{Maximum} & \textbf{Average} & \textbf{compression} \\
        \midrule
         AlexNet & 56 & 248 & 153.60 & 0.2875 & 0.5240 \\
         ResNet18 & 40 & 344 & 141.20 & 0.2242 & 0.6002 \\
         ResNet 34 & 64 & 312 & 151.78 & 0.2040 & 0.6890 \\
         ResNet50 & 64 & 576 & 124.08 & 0.2856 & 0.7018 \\
         ResNet101 & 64 & 560 & 125.38 & 0.4346 & 0.7256 \\
         InceptionV3 & 37 & 392 & 104.61 & 0.3411 & 0.7312 \\
         DenseNet121 & 32 & 356 & 60.80 & 0.1760 & 0.6896 \\
         VGG16 & 27 & 356 & 191.62 & 0.3449 & 0.6390 \\
         ConvNeXt & \zz8 & 224 & \zz21.40 & 0.4011 & 0.8017 \\
        \bottomrule
    \end{tabular}
    \end{adjustbox}
\end{table}

\subsection{Analysis of Results}\label{sec:anal}

During the training of the models, we record not only the accuracy but also the training time, but also
the number of epochs required for model convergence, and energy consumption. This additional 
information allows us to quantify the effects of compression on these models. In this section, 
we first review the total energy consumed during the retraining of these pretrained models, 
which tells us the energy required to achieve comparable performance after compression. 

We then investigate whether compressing a model leads to a shorter training time compared to the original model. 
In addition, we are interested in how many rounds of training the compressed models need to converge,
hence the number of epochs is provided. Finally, we examine the energy consumption for each round of training. 
Again, this analysis is based on training the compressed models as discussed in
Sections~\ref{sub:stego}, \ref{sub:prune}, and~\ref{sub:lr}, above.

\begin{figure}[!htb]
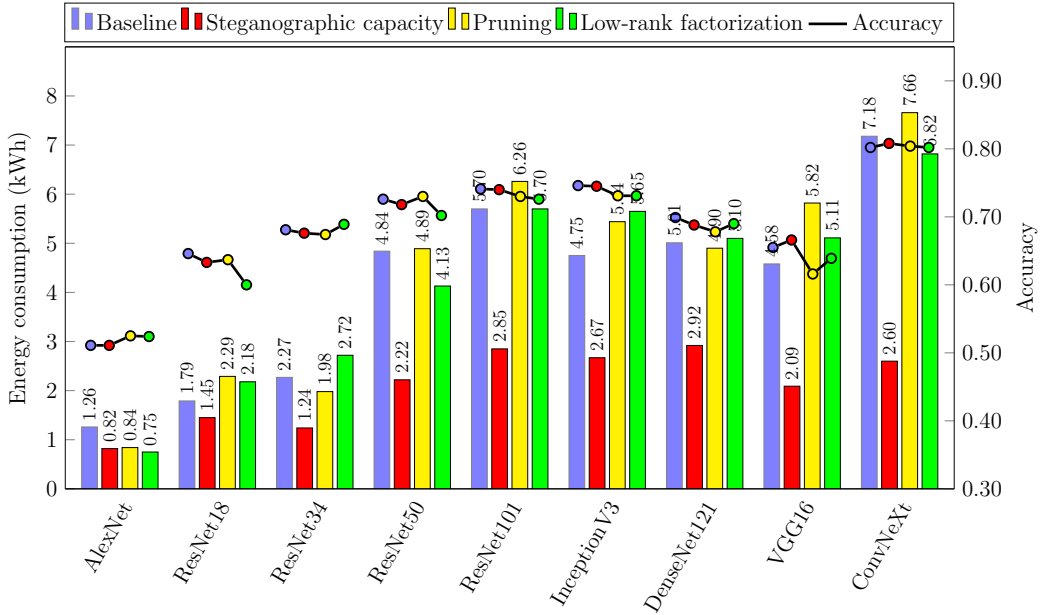

    \centering
    \begin{adjustbox}{scale=0.85}
    \input figures/energyAccuracy2.tex
    \end{adjustbox}
    \caption{Energy consumption and accuracy}
    \label{fig:enery_accuracy}
\end{figure}

Figure~\ref{fig:enery_accuracy} 
compares the energy consumption 
and accuracy achieved by each model under the different compression techniques. 
We observe that ConvNeXt is the most accurate model, with accuracies in
excess of~0.80 for the baseline case and all three compression techniques.   
In contrast, AlexNet performs the worst of the nine models tested,
with accuracies marginally above~0.50 for the~1000-class classification
problem used for our experiments.

From Figure~\ref{fig:enery_accuracy},
we see that pruning and low-rank factorization perform poorly
with respect to energy usage,
as models compressed using these techniques generally require a similar
amount of energy (if not more) to train than the corresponding uncompressed 
baseline models. The only notable exception is the poorest-performing
model, AlexNet, for which all three compression techniques perform similarly.
On the other hand, the steganographic capacity technique generally 
reduces the energy usage by a large margin. 

\begin{figure}[!htb]
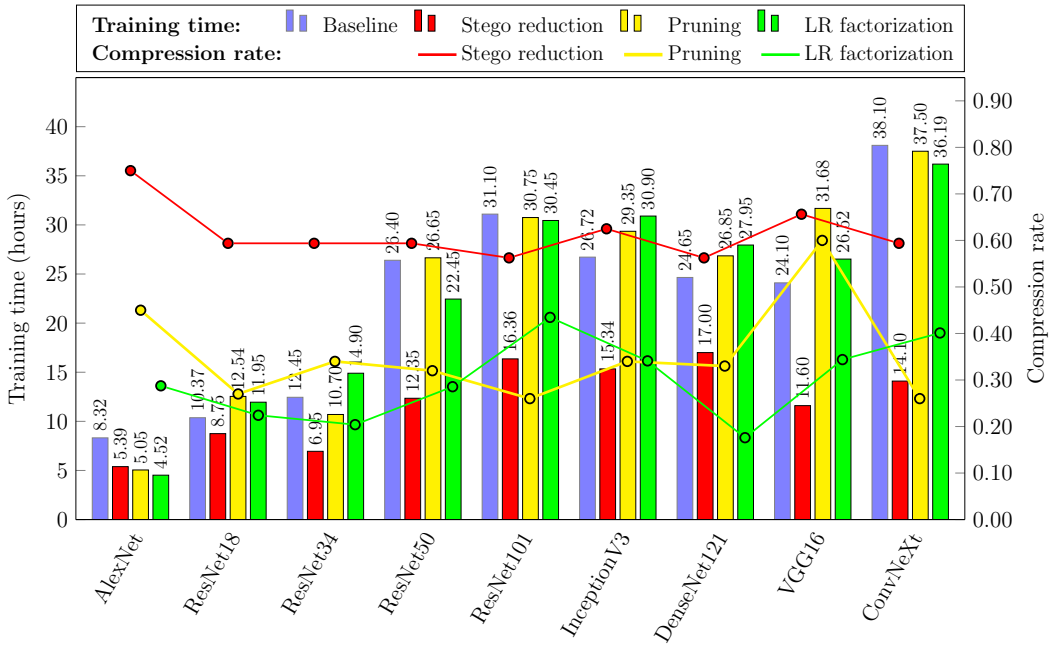

    \centering
    \begin{adjustbox}{scale=0.85}
    \input figures/timeCompress.tex
    \end{adjustbox}
    \caption{Training time and compression rate}
    \label{fig:time_compression}
\end{figure}

Figure~\ref{fig:time_compression} 
shows the training time required for each model 
and the corresponding compression rate. From the line graphs, we see that
steganographic capacity reduction provides the best compression results, while
the bar graphs show that steganographic capacity reduction
also performs best with respect to training times (the only exception
being AlexNet). In contrast, models compressed using either pruning or 
low-rank factorization often require more time to train than the corresponding
baseline models, while providing substantially less compression, as compared
to steganographic capacity reduction.
Again, by these measures, steganographic capacity reduction
is clearly the best approach.

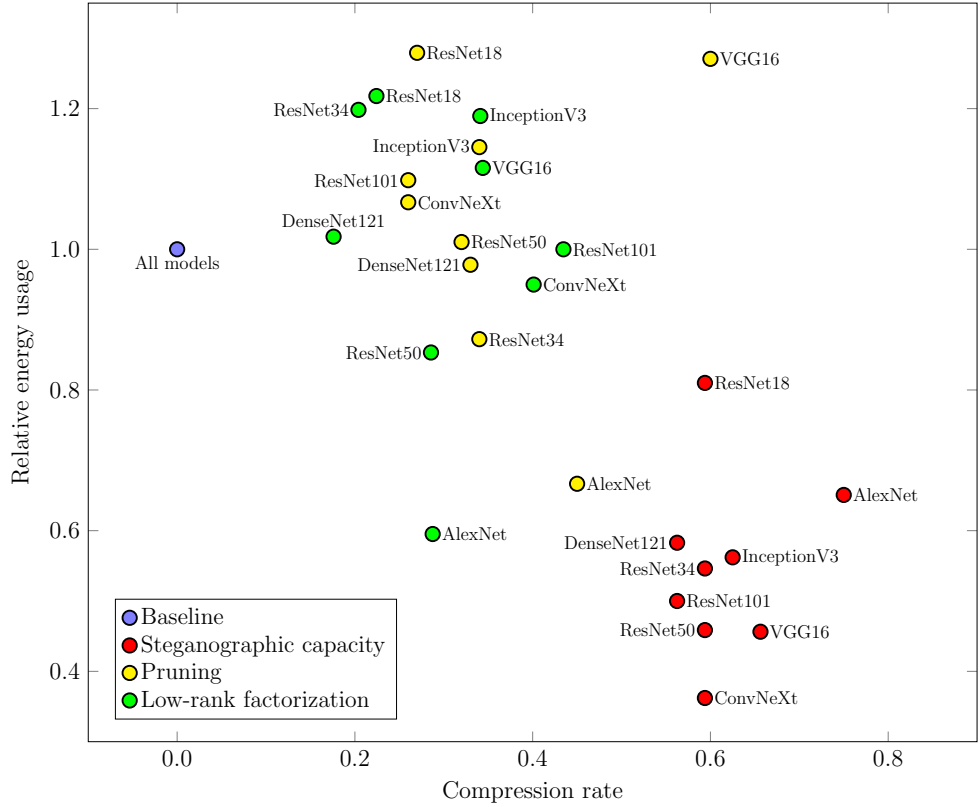
\begin{figure}[!htb]
    \centering
    \begin{adjustbox}{scale=0.90}
    \begin{tikzpicture}[scale=1.0, every node/.style={scale=1.0}]
\def\hh{\hspace*{0.025in}}
\pgfkeys{/pgf/number format/.cd,1000 sep={}}
\begin{axis}[
width  = 0.95\linewidth,
height = 12.5cm,
xmin=-0.10, xmax=0.90,
xtick={0.0, 0.2, 0.4, 0.6, 0.8},
ymin=0.3, ymax=1.35,
ytick={0.4, 0.6, 0.8, 1.0, 1.2},
xlabel style = {scale = 0.85},
ylabel style = {scale = 0.85},
x tick label style={scale = 0.85,
    	/pgf/number format/.cd,
   	fixed,
   	fixed zerofill,
    	precision=1},
y tick label style={scale = 0.85,
    	/pgf/number format/.cd,
   	fixed,
   	fixed zerofill,
    	precision=1},
nodes near coords,
every node near coord/.append style={scale=0.65, color=black, anchor=west, 
},
xlabel = {Compression rate}, 
ylabel = {Relative energy usage},
legend cell align=left,
legend pos=south west,
legend style={
	nodes={scale=0.825},
},
]
\addplot[ 
mark=*,
only marks,
mark size=3pt,
color=black,
thick,
text opacity=1.00,
fill=blue,
fill opacity=0.50,
point meta = explicit symbolic,
nodes near coords,
every node near coord/.append style={anchor=north}
]
coordinates {
(0.0,1.0) [All models]
};
\addlegendentry{Baseline}
\addplot[ 
mark=*,
mark size=3pt,
only marks,
color=black,
thick,
fill=red,
point meta = explicit symbolic,
nodes near coords,
]
coordinates {
(0.750000,0.650794) [\hh AlexNet]
(0.593800,0.810056) [\hh ResNet18]
(0.562500,0.500000) [\hh ResNet101]
(0.625000,0.562105) [\hh InceptionV3]
(0.656300,0.456332) [\hh VGG16]
(0.593800,0.362117) [\hh ConvNeXt]
};
\addlegendentry{Steganographic capacity}
\addplot[ 
forget plot,
mark=*,
mark size=3pt,
only marks,
color=black,
thick,
fill=red,
point meta = explicit symbolic,
nodes near coords,
every node near coord/.append style={anchor=east}
]
coordinates {
(0.593800,0.546256) [ResNet34\hh]
(0.593800,0.458678) [ResNet50\hh]
(0.562500,0.582834) [DenseNet121\hh]
};
\addplot[ 
mark=*,
only marks,
mark size=3pt,
color=black,
thick,
fill=yellow,
point meta = explicit symbolic,
nodes near coords,
]
coordinates {
(0.450000,0.666667) [\hh AlexNet]
(0.270000,1.279330) [\hh ResNet18]
(0.340000,0.872247) [\hh ResNet34]
(0.320000,1.010331) [\hh ResNet50]
(0.600000,1.270742) [\hh VGG16]
(0.260000,1.066852) [\hh ConvNeXt]
};
\addlegendentry{Pruning}
\addplot[ 
forget plot,
mark=*,
only marks,
mark size=3pt,
color=black,
thick,
fill=yellow,
point meta = explicit symbolic,
nodes near coords,
every node near coord/.append style={anchor=east}
]
coordinates {
(0.260000,1.098246) [ResNet101\hh]
(0.340000,1.145263) [InceptionV3\hh]
(0.330000,0.978044) [DenseNet121\hh]
};
\addplot[ 
mark=*,
only marks,
mark size=3pt,
color=black,
thick,
fill=green,
point meta = explicit symbolic,
nodes near coords,
]
coordinates {
(0.287500,0.595238) [\hh AlexNet]
(0.224200,1.217877) [\hh ResNet18]
(0.434600,1.000000) [\hh ResNet101]
(0.341100,1.189474) [\hh InceptionV3]
(0.343900,1.115721) [\hh VGG16]
(0.401100,0.949861) [\hh ConvNeXt]
};
\addlegendentry{Low-rank factorization}
\addplot[ 
forget plot,
mark=*,
only marks,
mark size=3pt,
color=black,
thick,
fill=green,
point meta = explicit symbolic,
nodes near coords,
every node near coord/.append style={anchor=south}
]
coordinates {
(0.176000,1.017964) [\raisebox{2pt}{DenseNet121}]
};
\addplot[ 
forget plot,
mark=*,
only marks,
mark size=3pt,
color=black,
thick,
fill=green,
point meta = explicit symbolic,
nodes near coords,
every node near coord/.append style={anchor=east}
]
coordinates {
(0.204000,1.198238) [ResNet34\hh]
(0.285600,0.853306) [ResNet50\hh]
};
\end{axis}
\end{tikzpicture}
    \end{adjustbox}
    \caption{Energy usage vs compression rate relative to baseline}
    \label{fig:energy_vs_compress}
\end{figure}

Figure~\ref{fig:energy_vs_compress}
correlates compression rates with energy consumption
relative to the corresponding baseline models. 
We observe that the steganographic capacity reduction
results are clustered in the lower-right corner, indicating
generally greater compression rates and lower energy consumption.
In fact, AlexNet is the only model for which steganographic
capacity reduction does not clearly outperform both
pruning and low-rank factorization. As noted above, AlexNet 
also yields the poorest accuracy of all of the models tested.

\begin{figure}[!htb]
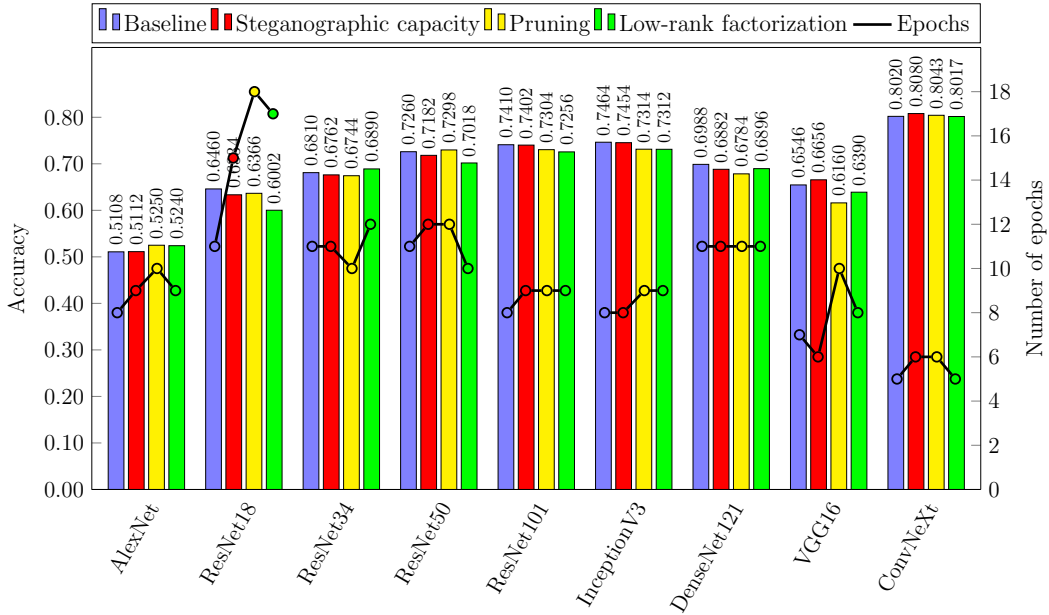

    \centering
    \begin{adjustbox}{scale=0.85}
    \input figures/epochsAccuracy.tex
    \end{adjustbox}
    \caption{Accuracy and number of epochs}
    \label{fig:accuracy_epoch}
\end{figure}

The bar graphs in Figure~\ref{fig:accuracy_epoch} represent the accuracy
of each model under each compression technique, while the line graphs
show the number of training epochs. We observe that compressed  
models generally require additional training epochs to achieve the
accuracy of the corresponding baseline model, and that pruning
tends to require the most training epochs.

\begin{figure}[!htb]
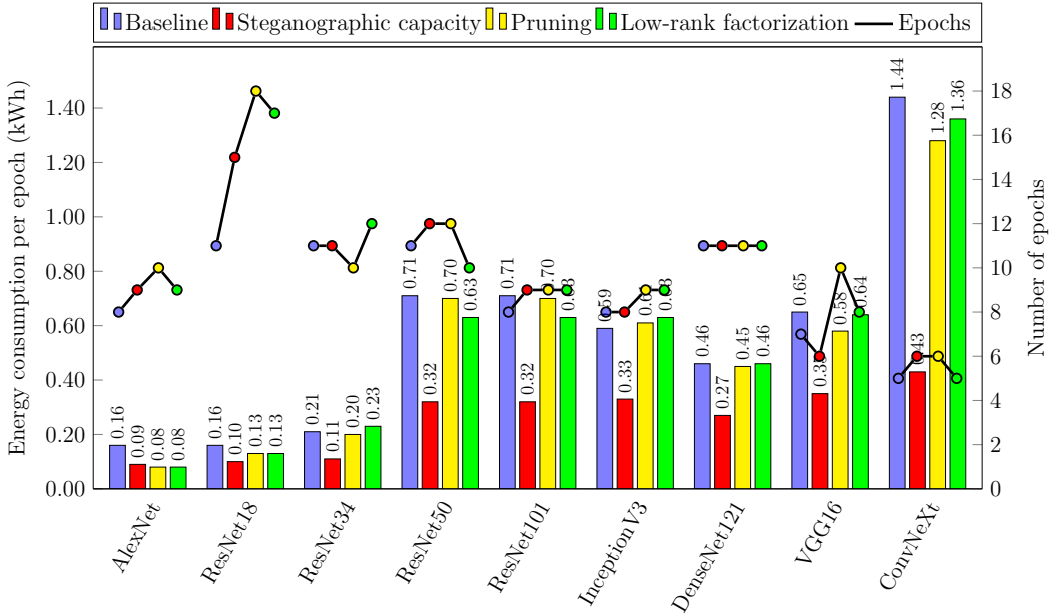

    \centering
    \begin{adjustbox}{scale=0.85}
    \input figures/epochsEnergy.tex
    \end{adjustbox}
    \caption{Energy consumption per epoch}
    \label{fig:energy_perepoch}
\end{figure}

The bar graphs in Figure~\ref{fig:energy_perepoch}  
represent the energy consumption per epoch.
As in Figure~\ref{fig:accuracy_epoch}, the line
graphs is Figure~\ref{fig:energy_perepoch} 
show the number of training epochs.
For every model except AlexNet, steganographic capacity reduction 
has the smallest energy usage per epoch. From this graph
and Figure~\ref{fig:enery_accuracy}, we deduce that even in cases
where the number of epochs increases under steganographic capacity,
the energy reduction per epoch is sufficient so that overall energy usage is 
reduced.

\subsection{Discussion}\label{sec:disc}

The results above show that steganographic capacity reduction effectively reduces 
energy usage by more than half for six of the models
(ResNet50, ResNet101, InceptionV3, DenseNet121, VGG16, and ConvNeXt).
On the other hand, neither pruning nor low-rank factorization
achieve consistent energy reductions
and, in fact, they result in increased energy usage in several cases.

Compressed models are generally able to achieve comparable accuracies after training. 
However, ResNet18 and VGG16 do suffer \hbox{significant} drops in accuracy for low-rank factorization
and pruning, respectively. None of the models suffer a loss in accuracy greater than~1\%\ under
steganographic capacity reduction. 

Steganographic capacity reduction also
significantly speeds up model training. Training 
ResNet50, ResNet101, InceptionV3, VGG16, and ConvNeXt using the 
steganographic capacity reduction technique takes approximately 
half the time needed to train the corresponding baseline models. 
In contrast, pruning and low-rank factorization fail to consistently reduce the
training time. 
With respect to the number of epoch, 
in some cases pruning dramatically increases
the number of required training epochs, which explains why pruning does not tend to
reduce training time or energy usage.

Based on our experimental results,
it is clear that reducing the size of model weights via quantization can 
greatly improve model training efficiency. However,
reducing the number of model weights via pruning or low-rank factorization 
does not produce any consistent positive impact on model training efficiency.

\section{Conclusions}\label{ch:conclusion}

Compression techniques, such as steganographic capacity reduction, 
pruning, and low-rank factorization, 
can be used to reduce the size of neural networks. 
While these methods are effective at making models more compact, we found in this study
that their effects on computational performance vary widely. Specifically, 
steganographic capacity reduction can provide a large reduction in energy 
usage and reduced training times, with only a minimal loss in accuracy.
However, pruning, and low-rank factorization appear to have little to offer
in terms of energy savings or improved training times.

For our steganographic capacity reduction experiments, we set
a maximum allowable loss in accuracy of~1\%. 
If we are willing to tolerate slightly larger losses in accuracy,
it is likely that substantial further reductions in training time
and energy usage could be achieved.
Trade-offs between training time, energy usage, and accuracy are 
especially critical for edge devices that have limited computational resources 
or power constraints. For example, a smart home hub may need to perform training 
on its own to immediately adapt to changes without offloading to the cloud. 
In such scenarios, it may be acceptable to sacrifice some accuracy in exchange 
for faster training or lower energy usage.

Further research on the combination of steganographic capacity reduction with other compression techniques 
would be worthwhile. Steganographic capacity reduction, which involved manipulation at the bit level, 
achieves a fairly consistent reduction in energy consumption across all model trainings. In contrast, 
pruning and low-rank factorization compress models at the level of neurons and weight matrices. 
Hence, combining steganographic capacity reduction with these techniques is technically viable 
and could potentially achieve a higher model compression rate and lower energy consumption.

Another area for further research would be to investigate whether compression is also an effective 
strategy for Large Language Models (LLMs), particularly from the perspective of energy conservation.
Given the rapid growth of LLMs, in terms of model size and depth, 
it is anticipated that the associated electricity usages for training such models
will increase at a similar rate~\cite{Samsi_2023}. Therefore, investigating whether compression 
can effectively reduce energy consumption of such models would be worthwhile. 
This also aligns with the broader goal of sustainable AI practices, which can
ensure that advances in AI are not achieved at the expense of the environment.

\bibliographystyle{plain}
\bibliography{references}

\end{document}